\pdfoutput=1

\documentclass[11pt]{article}

\usepackage[dvipsnames]{xcolor}
\usepackage[]{acl}

\usepackage{times}
\usepackage{latexsym}

\usepackage[T1]{fontenc}

\usepackage[utf8]{inputenc}

\usepackage{microtype}

\usepackage{inconsolata}

\usepackage{graphics}
\usepackage{multirow}
\usepackage{fancyhdr}
\usepackage{subcaption}
\usepackage{booktabs}

\newcommand{\eg}{\textit{e.g.}}
\newcommand{\ie}{\textit{i.e.}}

\title{The CLC-UKET Dataset: Benchmarking Case Outcome Prediction for the UK Employment Tribunal}

\author{Huiyuan Xie\textsuperscript{\ensuremath{1}} \hspace{0.5em} Felix Steffek\textsuperscript{\ensuremath{2}} \hspace{0.5em} Joana Ribeiro de Faria\textsuperscript{\ensuremath{2}} \\
\textbf{Christine Carter}\textsuperscript{\ensuremath{2}} \hspace{0.5em} \textbf{Jonathan Rutherford}\textsuperscript{\ensuremath{2}} \\
\textsuperscript{\ensuremath{1}}Department of Computer Science and Technology, University of Cambridge 
\\ \textsuperscript{\ensuremath{2}}Faculty of Law, University of Cambridge
}

\begin{document}

\maketitle

\begin{abstract}

This paper explores the intersection of technological innovation and access to justice by developing a benchmark for predicting case outcomes in the UK Employment Tribunal (UKET). To address the challenge of extensive manual annotation, the study employs a large language model (LLM) for automatic annotation, resulting in the creation of the CLC-UKET dataset. The dataset consists of approximately 19,000 UKET cases and their metadata. Comprehensive legal annotations cover facts, claims, precedent references, statutory references, case outcomes, reasons and jurisdiction codes. Facilitated by the CLC-UKET data, we examine a multi-class case outcome prediction task in the UKET. Human predictions are collected to establish a performance reference for model comparison. Empirical results from baseline models indicate that finetuned transformer models outperform zero-shot and few-shot LLMs on the UKET prediction task. The performance of zero-shot LLMs can be enhanced by integrating task-related information into few-shot examples. We hope that the CLC-UKET dataset, along with human annotations and empirical findings, can serve as a valuable benchmark for employment-related dispute resolution.
\end{abstract}

\section{Introduction}


In recent years, there has been great interest in adopting natural language processing techniques in the legal domain. One notable application is the prediction of outcomes for legal disputes in various jurisdictions \citep{xiao2018cail,poudyal2020echr,hwang2022multi,henderson2022pile,niklaus2023lextreme}. 
However, the AI-based prediction of UK court decisions is still under-explored. 

This paper investigates the prediction of dispute outcomes in the UK Employment Tribunal (UKET). The UKET serves a crucial function in the UK justice system, specifically dealing with employment-related disputes. Cases heard at the UKET cover a wide range of issues, such as unfair dismissal, discrimination and breach of contract. The possibility to apply to the UKET for a decision ensures that employment rights can be enforced. Knowing the likely outcome of a court procedure improves access to justice and facilitates amicable dispute resolution. 

The contributions of this paper are as follows:
\begin{enumerate}
    \item We constructed a large-scale CLC-UKET dataset based on the Cambridge Law Corpus (CLC) \citep{ostling2023cambridge}. CLC-UKET includes two components: CLC-UKET$_{anno}$ and CLC-UKET$_{pred}$. CLC-UKET$_{anno}$ consists of a selection of 19,090 UKET case judgments heard between 2011 and 2023 (inclusive). All cases come with metadata including a unique case identifier, the hearing date and jurisdiction codes. We further provided detailed legal annotations for all cases, including (a) facts, (b) claims, (c) references to legal statutes, acts, regulations, provisions and rules, (d) references to precedents and other court decisions, (e) general case outcome and (f) detailed order and remedies. We further curated CLC-UKET$_{pred}$, specifically designed to facilitate a multi-class case outcome prediction task. CLC-UKET$_{pred}$ consists of 14,582 cases, each supplemented with statements detailing the facts, claims and the general outcomes of the cases. 
    \item We assessed human performance on the UKET outcome prediction task on CLC-UKET$_{pred}$ with the aim of setting a human performance reference to calibrate prediction models. 
    \item We experimented with a range of baseline models to predict the general case outcomes based on information about facts and claims of UKET cases.
\end{enumerate}

The CLC-UKET dataset and the empirical explorations aim to supplement the standard CLC dataset and facilitate future research on employment-related dispute resolution in the UK legal system. We will make the CLC-UKET dataset available via the official CLC website\footnote{The CLC website: https://www.cst.cam.ac.uk/research/srg/\\
projects/law.}.

\section{UK Legal System and UKET}

The UK has a special category of judicial body, the employment tribunals, which deal exclusively with employment disputes. The UKET is one of the three largest tribunals in the greater tribunals system \citep{Office}. The UKET aims to provide a procedure which is easily accessible, informal, speedy and inexpensive \citep[p. 23]{Beis}.
The form of employment tribunal proceedings is adversarial rather than investigatory, as each party has to present and prove its case \citep{Deakin}. Claimants must comply with procedural and substantive requirements to be successful. For instance, claimants must submit their claims on time, comply with the orders of the tribunal, present required evidence or information in a timely manner, and avoid scandalous, unreasonable or vexatious conduct (which would make a fair trial impossible). These are usually considered as \textit{procedural} requirements. Claimants must also comply with the substantive requirements of the rules supporting the claims. For example, in order to be successful with a discrimination claim on grounds of disability, the claimant must prove their status as an employee, demonstrate their disability and show that they faced discrimination, which are considered as \textit{substantive} requirements of the case. These procedural and substantive requirements are not necessarily determined at one final hearing or included in one final judgment. Instead, they may be iteratively decided at different stages, which can result in multiple decisions. 

The employee (claimant) and the employer (respondent) submit their claims and responses, respectively, through a standardised form (Rules 8 and 16 of the Employment Tribunals Rules of Procedure 2013, hereinafter referred to as \citet{Rules}). The tribunal considers these forms and may dismiss a claim for procedural or substantive reasons, \eg, for lack of jurisdiction or for lack of any reasonable prospect of success \citep[r. 27]{Rules}. At any stage of the proceedings, the tribunal can determine a preliminary issue, make a procedural order (\eg, a deposit order or require the presentation of additional documents) or make a final decision (\eg, strike out the claim, \citet[r. 37]{Rules}). 
There may be multiple final hearings for different issues, for example, one hearing to determine whether a party is liable, another hearing to determine the remedy and another to determine the costs \citep[r. 57]{Rules}. Each of these hearings results in a separate judgment, written out in a separate document. Finally, a party may request a reconsideration of a previous judgment, which will lead to another judgment \citep[r. 70]{Rules}. As a consequence, the resolution of a dispute may not be covered by one judgment only, but may be determined by iterative multiple decisions resulting in various case documents.

Each decision is linked to one or multiple jurisdiction codes. 
In the case of the UKET, there are 54 jurisdiction codes in total, which are used to identify the matter of disputes. 
By way of example, the jurisdiction code ``unfair dismissal'' is used when claimants argue that they have been unfairly dismissed. 
This jurisdiction code is often used in addition to other jurisdiction codes, such as unlawful deduction from wages, redundancy, breach of contract and working time regulations. 

In stark contrast with typical UK judgments, UKET decisions are relatively clearly structured, not only because there are no dissenting opinions, but also because there are specific rulings that set out which elements a judgment must contain \citep[r. 62(5)]{Rules}. Nevertheless, UKET judgments are not always consistent since there are no formal rules on the style to be used in drafting a decision. Most English judgments summarise their decisions in a paragraph, although this summary does not need to respect any particular form \citep[p. 136]{Conseil}. In the case of the UKET, the summary is often found at the beginning of the judgment. However, judgments on multiple claims are sometimes divided into chapters, each analysing one claim containing the relevant decision. Also, whilst a judgment may contain an initial statement that the claimant is successful, it may not be clear which claim(s) this relates to in cases where there are multiple claims. 

\section{Related Work}

\subsection{Analysis of Employment Judgments}
Quantitative methods for analysing legal judgments have long been explored. In relation to employment law, \citet{Grunbaum} analysed 20 US Supreme Court judgments to identify the variables which impacted outcomes. Similarly, \citet{Field} identified factors which influenced outcomes of performance appraisal judgments. \citet{Brudney} analysed the extent to which extradoctrinal factors such as political party, gender and professional experience influenced outcomes. 

Moreover, several studies explored correlations between specific demographic groups and the ability to pursue their employment rights in tribunal. In the US, \citet{Schuster} analysed 153 federal court cases, focusing on age discrimination, whilst \citet{Schultz} investigated race and sex discrimination. In the UK, \citet{Barnard} investigated whether EU-8 migrant workers were able to enforce their rights by bringing claims before the UKET. 

Many of these studies occurred before judgments were published online, and therefore not only entailed costly journeys to the registers, but also required manual extraction and tagging of specific elements of court decisions. More recently, \citet{Blackham} conducted quantitative analyses of employment decisions, but despite having access to online judgments, some of their tasks still required manual labour.

%

\subsection{Legal Judgment Prediction}

The advance of deep learning models alongside the development of large-scale legal datasets has greatly advanced the research on legal judgment prediction (LJP) \citep{xiao2018cail,chalkidis2019neural,osullivan2019predicting,ma2021legal_msjudge,chalkidis2023lexfiles, Colombo}. A large number of datasets have been created for both civil law systems \citep{poudyal2020echr,yamada-etal-2022-annotation} and common law systems \citep{cap2018caselaw,henderson2022pile,ostling2023cambridge,butler2024how_australian_corpus}.

Facilitated by large-scale datasets, there has been a surge in the application of deep learning models to LJP in recent years. \citet{zhong-etal-2018-legal} introduced TopJudge to address LJP using multi-task learning that combines three aspects: law articles, charges and terms of penalty. Another notable contribution is the work of \citet{ma2021legal_msjudge} where an end-to-end framework was built to predict dispute outcomes using multi-task supervision and multi-stage representation learning. To the best of our knowledge, the only notable LJP paper on UK law is \citet{strickson2020legal}, which dates before the emergence of LLMs and is limited to the binary task of UK Supreme Court judges allowing or dismissing an appeal.



\section{The CLC-UKET Dataset}

We curated a large-scale dataset focusing on UK employment-related dispute resolution. The resulting CLC-UKET dataset consists of two components: CLC-UKET$_{anno}$ consisting of 19,090 cases with detailed legal annotations and CLC-UKET$_{pred}$ with 14,582 cases curated for case outcome prediction for the UKET. The CLC-UKET dataset is constructed based on the UKET subset of the CLC \citep{ostling2023cambridge} by adding annotations for selected UKET cases. A common practice for collecting legal annotations is to ask legal experts to manually annotate texts. However, this can be costly and time-consuming. To alleviate the burden of manual annotation, we explored utilising large language models (LLMs) for automatic annotation. 

The dataset curation pipeline of CLC-UKET$_{anno}$ consists of two steps: a \textit{case preparation} module and an \textit{LLM-aided case annotation} module. 

\subsection{Case Preparation}

The raw UKET subset of the CLC contains 52,339 cases in total, covering employment-related cases heard at the UKET from January 2011 to August 2023 (inclusive).\footnote{The hearing dates of the cases in the UKET subset of the CLC range from 2011 to 2023, although some cases were submitted before 2011.} After analysing these cases, we noticed that many cases only consist of one page as regards the tribunal decision. Based on the observations from \citet{defaria2024automatic}, many of these cases involve straightforward procedural decisions, for example when claimants withdraw their cases or respondents do not respond at all such that a default judgment is made. As such cases do not contain substantial information on facts and substantive reasons, we excluded them at the case preparation step.

After this filtering step, we obtained a collection of 19,090 cases containing more than one page in their court decision files. For each case, we collected a list of metadata, including a unique case identifier used in the UKET records, date of filing, date of decision, place of hearing, judges, claimant(s), respondent(s) and appearances at the hearing. We also obtained jurisdiction codes for all cases from the UKET website.\footnote{https://www.gov.uk/employment-tribunal-decisions.} In legal contexts, a \textit{jurisdiction code} typically refers to a numerical or alphanumeric code assigned to a specific legal jurisdiction, a certain subject matter or a geographic area. Legal jurisdictions are defined areas with a distinct set of laws and regulations. In the case of the UKET, there are a total of 54 jurisdiction codes, which are used to identify the dispute matter. Each UKET case can be associated to multiple jurisdiction codes that indicate the categorical areas of the case. As an example, the code \textit{unfair dismissal} is used when claimants argue that they have been unfairly dismissed and submit a claim for payment of a certain sum, \eg, basic award, compensatory award (lack of notice pay and loss of earnings between a period) and injury to feelings award. 
A full list of jurisdiction codes in the UKET is presented in Appendix \ref{app:juris_codes}.

\subsection{LLM-aided Case Annotation}
\label{subsec:llm_annotation}
The CLC provides raw texts of the decisions of UKET cases. These documents usually contain entangled statements about facts provided by parties and their lawyers, reasoning towards a decision, legal statutes and precedents applied to justify the reasoning and final decisions regarding the case outcome. In this step, we followed similar lines to \citet{defaria2024automatic} and utilised the GPT-4-turbo model \citep{achiam2023gpt} to automatically extract legal information from UKET decisions. 

We applied an iterative development process to find the optimal prompt for the purpose of legal information extraction. The final prompt that we opted for yielded the best results in terms of the accuracy of information extracted, the adequacy of necessary information contained therein and the level of detail. The final prompt that we used for LLM-aided case annotation is presented in Appendix \ref{app:prompts}. After automatic annotation, we obtained detailed annotations on important legal factors for 19,090 CLC-UKET cases, covering (1) facts, (2) claims, (3) references to legal statutes, (4) references to precedents, (5) general case outcomes, (6) general case outcomes labelled as ``claimant wins'', ``claimant loses'', ``claimant partly wins'' and ``other'', (7) detailed orders and remedies and (8) reasons.

\section{Case Outcome Prediction}

The annotated CLC-UKET data (\ie, CLC-UKET$_{anno}$) provides a large collection of court decisions augmented with rich legal annotations, which can readily be used for downstream legal AI tasks. In this paper, we showcase a use case of the CLC-UKET data by investigating a classic task in legal AI, i.e., case outcome prediction.

\subsection{Task Definition}

Given a set of facts and claims of a UKET case, the task of case outcome prediction aims to automatically generate an outcome label falling into one of the following four categories: ``claimant wins'', ``claimant loses'', ``claimant partly wins'' and ``other''. The facts and the claims are the judges' summarisation of the statements provided by the claimant(s) and respondent(s) prior to or during a hearing.

More formally, given a set of facts $F = f_1, f_2, \cdots, f_m$ and a set of claims $C = c_1, c_2, \cdots, c_n$ for a UKET case, a prediction model \texttt{CLS} outputs a label $g$ for the general case outcome:

\[
g = \texttt{CLS}(F, C)
\]

where $g \in$ [``claimant wins'', ``claimant loses'', ``claimant partly wins'' and ``other''].

Note that there is a debate concerning the difference between the legal judgment prediction (LJP) task and the case outcome classification (COC) task \citep{medvedeva2021automatic,santosh2022deconfounding,medvedeva2023rethinking,medvedeva2023legal}. In this paper, we opted for the terminology ``prediction'' over ``classification'' as we deliberately excluded explicit information about case outcomes from the input of the prediction task, and only kept descriptions of facts and claims in the input. As such, this task focuses on predicting case outcomes based solely on information about facts and claims. Similarly, the legal experts predicting outcomes had only access to facts and claims.

\subsection{Data Preparation for the Prediction Task}

We tailored the CLC-UKET$_{anno}$ data to construct a case outcome prediction task for the UKET. Three types of legal factors are needed for the prediction task, namely \textit{facts}, \textit{claims} and \textit{general case outcomes}. The input to the prediction models is a sequence of fact statements concatenated with claim statements, in the form of ``\textit{fact}$_1$, \textit{fact}$_2$, $\cdots$, \textit{fact}$_n$ [SEP] \textit{claim}$_1$, \textit{claim}$_2$, $\cdots$, \textit{claim}$_m$''. The target output of the prediction task is a general outcome label, which is a categorical variable labelling potential case outcomes as \textit{claimant wins}, \textit{claimant partly wins}, \textit{claimant loses} and \textit{other}. 

\begin{table}[t]
    \centering
    \resizebox{0.8\columnwidth}{!}{
    \begin{tabular}{c|ccc}
    \hline
    & \textbf{train} & \textbf{val} & \textbf{test} \\ \hline
    \#Cases & 11,838 & 1,373 & 1,371 \\
    \#AvgFactLen & 79 & 85 & 88 \\
    \#MaxFactLen & 409 & 463 & 321 \\
    \#AvgClaimLen & 34 & 34 & 34 \\
    \#MaxClaimLen & 187 & 164 & 150 \\\hline
    \end{tabular}
    }
    \caption{Data statistics of the CLC-UKET$_{pred}$ dataset. \#Cases denotes the number of cases. \#AvgFactLen denotes the average number of words per fact statement. \#MaxFactLen denotes the maximum length of fact statements. \#AvgClaimLen denotes the average number of words per claim statement. \#MaxClaimLen denotes the maximum length of claim statements.}
    \label{tab:data_stats}
\end{table}

\subsection{Data Statistics}
From the 19,090 cases in the CLC-UKET$_{anno}$ dataset, we filtered out cases where no substantial information about \textit{facts} and \textit{claims} was extracted by GPT-4 at the LLM-aided case annotation step. After the filtering, we obtained a set of 14,582 UKET cases, supplemented with fact and claim statements extracted by GPT-4. We denote this prediction dataset as CLC-UKET$_{pred}$. Following general practice in machine learning research, we divided the 14,582 CLC-UKET$_{pred}$ cases into three splits: \textit{training}, \textit{validation} and \textit{testing}. The details on data statistics of the train/val/test sets for CLC-UKET$_{pred}$ are summarised in Table \ref{tab:data_stats}. 

Note that for the \textit{training} and \textit{validation} sets, all three legal factors - facts, claims and outcomes - were sourced from information automatically extracted by GPT-4, as detailed in Section \ref{subsec:llm_annotation}. For the \textit{testing} set, facts and claims were automatically extracted by GPT-4, whilst the case outcome labels were manually annotated by a legal expert\footnote{The legal annotator is a PhD Candidate in Law.}. The expert annotator carefully analysed the full court judgments and summarised the judges' decisions into general case outcome labels. These manually annotated outcome labels for the test cases represent the actual judicial decisions, serving as gold-standard references for prediction evaluation.

\section{Experiments and Results}

\subsection{Baseline Models}

We experimented with two classes of baseline models:

\begin{enumerate}
    \item Transformer-based \citep{vaswani2017attention_transformer} models, including BERT \citep{devlin2019bert} and T5 \citep{raffel2020exploring_t5};
    \item LLM-based models, including GPT-3.5 \citep{openai2022introducing} and GPT-4 \citep{achiam2023gpt}. 
\end{enumerate}

The two Transformer-based models were fine-tuned on our CLC-UKET$_{pred}$ data, whilst GPT-3.5 and GPT-4 were tested using zero-shot and few-shot settings without dedicated fine-tuning. Implementation details of the baseline models are presented in Appendix \ref{app:experiment_details}.


\textbf{BERT}. 
We fine-tuned BERT \citep{devlin2019bert} on the training set of CLC-UKET$_{pred}$ with the Adam optimiser \citep{kingma2014adam} with a learning rate of 1e-4 and a batch size of 32. The final checkpoint was obtained after training the model for 5 epochs. 

\textbf{T5}. The T5 model \citep{raffel2020exploring_t5} is also fine-tuned on the training set of CLC-UKET$_{pred}$. The model is optimised with a learning rate of 1e-4 for 5 epochs. 

\textbf{GPT-3.5-turbo} and \textbf{GPT-4-turbo}. We tested GPT-based models with diverse settings, including (1) zero-shot prediction, (2) few-shot prediction with randomly selected examples and (3) few-shot prediction with examples selected according to jurisdiction codes. The prompts that we used for LLM experiments are presented in Appendix \ref{app:prompts}.

\begin{itemize}
    \item \textit{Zero-shot prediction}. In this setting, the GPT-based models are directly asked to predict an outcome based on information about facts and claims of a case. No examples are provided to the models in the prompts. 
    \item \textit{Few-shot prediction with randomly selected examples}. We randomly selected a few examples from the training set and included them in the prompt to GPT-based models. We also investigated the effects of the number of examples on prediction performance by experimenting with two numbers (\ie, 2 and 5) for examples included in the prompts. 
    \item \textit{Few-shot prediction with examples selected using jurisdiction codes}. This setting differs from the above few-shot setting in that we deliberately sampled case examples according to jurisdiction code similarity. In other words, given a target case for which a case outcome is to be predicted, we first identified the set of jurisdiction codes associated with it. Next, we gathered a collection of cases that share at least one jurisdiction code with the target case. From this collection, we sampled a specified number (similarly, 2 and 5) of example cases to include in the few-shot prompt.
\end{itemize}

\subsection{Human Prediction}
\label{sec:human_pred}
We further investigated how well legal experts can predict UKET case outcomes given facts and claims. This investigation is of paramount importance, as human performance can establish a reference to calibrate model performance.

Two legal experts conducted the human prediction exercise. They are PhD candidates in Law with a focus on UK employment law. They were supervised by a professor of law based in the UK. Each test case in CLC-UKET$_{pred}$ was separately annotated by the two legal experts. We asked annotators to indicate what they think is the most likely case outcome after reading facts and claims of a case. They were also asked to indicate whether a prediction is of low confidence. Cases labelled with low confidence are usually cases that are hard to predict due to insufficient information contained in the given facts or claims or due to the intrinsic complexity of a case (in particular the claims raised). 

At the beginning of the annotation process, both annotators were provided with annotation guidelines (see Appendix \ref{app:anno_guidelines} for details). The annotation guidelines are consistent with our overarching experimentation design for the prediction task. Annotators were asked to make their judgments separately, avoiding discussions amongst themselves. We emphasised that human predictions should be made based on the same facts and claims that prediction models were evaluated on. Annotators were required not to search for the cases they were annotating on the internet.\footnote{However, annotators were free to research other information that might be helpful for the annotations, for example, information on the applicable law.} Whenever questions regarding the implementation of the annotation arose during the annotation process, the annotators were provided with clarification by the supervisor.

After annotating, we obtained two independent sets of predicted case outcome labels for the 1,371 test cases. The Cohen's Kappa score for all annotations is 0.421\footnote{The Cohen's Kappa score between two specialised legal experts ranges from 0.41 to 0.60 indicating moderate agreement, highlighting the inherent complexity in the UKET prediction task.}.

\subsection{Results} \label{subsec:results}

\begin{table}[t]
    \centering
    \resizebox{\columnwidth}{!}{
    \begin{tabular}{c|cccc}
    \hline
        \textbf{Baseline} & \textbf{Accuracy} & \textbf{Precision} & \textbf{Recall} & \textbf{F-score} \\ \hline
        Random & 0.241 & 0.340 & 0.241 & 0.276 \\ 
        BERT & 0.446 & \textbf{0.623} & 0.446 & 0.427 \\
        T5 & \textbf{0.624} & 0.602 & \textbf{0.624} & \textbf{0.564} \\ 
        GPT-3.5$_{zero}$ & 0.535 & 0.553 & 0.535 & 0.525 \\
        GPT-3.5$_{rand2}$ & 0.540 & 0.567 & 0.540 & 0.535 \\
        GPT-3.5$_{rand5}$ & 0.532 & 0.561 & 0.532 & 0.532 \\
        GPT-3.5$_{juris2}$ & 0.544 & 0.568 & 0.544 & 0.542 \\
        GPT-3.5$_{juris5}$ & 0.549 & 0.570 & 0.549 & 0.550 \\
        GPT-4$_{zero}$ & 0.545 & \textbf{0.623} & 0.545 & 0.549 \\
        GPT-4$_{rand2}$ &0.518 & 0.612 & 0.518 & 0.530 \\
        GPT-4$_{rand5}$ & 0.539 & 0.614 & 0.539 & 0.547 \\
        GPT-4$_{juris2}$ & 0.540 & 0.619 & 0.540 & 0.551 \\
        GPT-4$_{juris5}$ & 0.536 & 0.617 & 0.536 & 0.546 \\\hline\hline
        \textbf{Human} &0.693 & 0.680 & 0.693 & 0.672 \\ \hline
    \end{tabular}
    }
    \caption{Overall evaluation results for the multi-class CLC-UKET$_{pred}$ prediction task. \textit{Precision}, \textit{recall} and \textit{F-score} report the weighted average of precision/recall/F-score obtained across labels, accounting for label imbalance. \textit{Random} refers to \textit{random guess}. \textit{Human} refers to the averaged scores of the outcome labels predicted by two human experts. All predicted outcomes were evaluated against gold-standard case outcome labels directly extracted from court decisions.}
    \label{tab:results}
\end{table}

\textbf{Overall results}. Table \ref{tab:results} presents the overall evaluation results for the CLC-UKET$_{pred}$ prediction task. The experiment findings reveal several key insights regarding the performance of different models. All models tested significantly outperform the random guess baseline, indicating the models' efficacy on this task. Among the models, the fine-tuned T5 emerges as the best performer overall, achieving the highest F-score. There is a noticeable gap between machine and human performance, with human expert predictions obtaining a 19.1\% higher F-score compared to the fine-tuned T5, highlighting the superiority of human judgment in this domain in a baseline setting. 

In terms of the two GPT-based models, GPT-4 generally outperforms GPT-3.5, reinforcing the advancements made in this newer model version. However, the margin of GPT-4's outperformance is rather small. The inclusion of few-shot examples proves beneficial for improving GPT-3.5’s prediction performance. Specifically, using examples that share similar jurisdiction codes with the target case enhances the F-score of GPT-3.5’s predictions more effectively than randomly sampled examples, validating the positive impact of incorporating task-specific information on GPT-3.5's prediction performance. In addition, the marginal gains observed when varying the number of few-shot examples provided to GPT-based models suggest that simply increasing the number of examples is not sufficient to significantly boost performance. Moreover, GPT-4, in its zero-shot setting, already achieves the highest precision among all baseline models. Providing two similar cases in the \textit{juris-2} few-shot setting improves GPT-4's F-score compared to the zero-shot setting.

\begin{table}[t]
    \centering
    \resizebox{0.95\linewidth}{!}{
    \begin{tabular}{c|c|ccc}
    \hline
         \textbf{Baseline} & \textbf{Label} & \textbf{Precision} & \textbf{Recall} & \textbf{F-score} \\ \hline
         \multirow{4}{*}{BERT} & wins & 0.459 & 0.828 & 0.591 \\
         & loses & 0.869 & 0.215 & 0.345 \\
         & partly & 0.381 & 0.364 & 0.372 \\        
         & other & 0.036 & 0.455 & 0.067 \\
         \hline
         \multirow{4}{*}{T5} & wins & 0.595 & 0.716 & 0.650 \\
         & loses & 0.647 & 0.846 & 0.734 \\
         & partly & 0.541 & 0.066 & 0.117 \\
         & other & 0 & 0 & 0 \\ \hline
         \multirow{4}{*}{GPT-3.5$_{juris5}$} & wins & 0.515 & 0.700 & 0.594 \\
         & loses & 0.720 & 0.565 & 0.633 \\
         & partly & 0.362 & 0.305 & 0.331 \\
         & other & 0.143 & 0.455 & 0.217 \\ \hline
         \multirow{4}{*}{GPT-4$_{juris2}$} & wins & 0.588 & 0.700 & 0.639 \\
         & loses & 0.778 & 0.430 & 0.554 \\
         & partly & 0.359 & 0.541 & 0.431 \\
         & other & 0.082 & 0.364 & 0.133 \\ \hline\hline 
         \multirow{4}{*}{\textbf{Human}} & wins & 0.627 & 0.815 & 0.708 \\
         & loses & 0.792 & 0.812 & 0.802 \\
         & partly & 0.554 & 0.302 & 0.391 \\
         & other & 0 & 0 & 0 \\
         \hline
    \end{tabular}
    }
    \caption{Evaluation scores obtained by baseline models and human predictions for the four label categories: \textit{claimant wins}, \textit{claimant loses}, \textit{claimant partly wins} and \textit{other}. The numbers of cases for the four labels are 437, 618, 305 and 11, respectively. For GPT-3.5 and GPT-4, the variants that achieved the highest F-sclores across relevant settings are presented.}
    \label{tab:individual_scores}
\end{table}

\textbf{Results for individual classes}. In Table \ref{tab:individual_scores} we report the individual scores achieved by baseline models and human predictions across various label categories. Most baseline models demonstrate a high recall and a relatively low precision when predicting ``claimant wins'' and in contrast achieve a high precision and a relatively low recall when predicting the ``claimant loses'' label. These findings underscore the distinct trade-offs that prediction models make between precision and recall. Human predictions exhibit strong performance in the ``claimant wins'' and ``claimant loses'' categories, where the F-scores are consistently high. The labels ``partly wins'' and  ``other'' consistently receive lower evaluation scores across all models and the human predictors, which may be attributed to the inherent difficulty of identifying cases within these two categories, compounded by the imbalanced distribution of cases across four categories. 


\begin{table}[t]
    \centering
    \resizebox{\columnwidth}{!}{
    \begin{tabular}{c|cccc}
    \hline
        \textbf{Baseline} & \textbf{Accuracy} & \textbf{Precision} & \textbf{Recall} & \textbf{F-score} \\ \hline
        BERT & 0.443 & \textbf{0.619} & 0.443 & 0.421 \\
        T5 & \textbf{0.535} & 0.552 & \textbf{0.535} & \textbf{0.480} \\
        GPT-3.5$_{juris5}$ & 0.455 & 0.488 & 0.455 & 0.451 \\
        GPT-4$_{juris2}$ & 0.465 & 0.527 & 0.465 & 0.448 \\ \hline\hline
        \textbf{Human} & 0.477 & 0.507 & 0.477 & 0.448 \\ \hline
    \end{tabular}
    }
    \caption{Evaluation results obtained by baseline models and human predictions on test cases which are considered as hard to predict by human experts.}
    \label{tab:low_confidence}
\end{table}

\textbf{Performance on \textit{low confidence} cases}. In the human prediction process described in Section \ref{sec:human_pred}, expert annotators were asked to explicitly indicate whether a case was difficult for them to predict based on the given facts and claims (\ie, a ``low confidence'' prediction). Using these annotations, we further analysed different baselines for cases that were considered difficult by the human experts. Comparing Table \ref{tab:low_confidence} with Table \ref{tab:results}, it can be observed that human performance on predicting for the \textit{low confidence} cases is significantly worse than for all cases, suggesting that human assessments of the difficulty level of the prediction task align well with the empirical results. Furthermore, all baseline models exhibit relatively lower scores when evaluated on the \textit{low confidence} cases. This pattern indicates that cases that are more challenging for human experts are also more difficult for the models.

\begin{table}[t]
    \centering
    \resizebox{\columnwidth}{!}{
    \begin{tabular}{c|cccc}
    \hline
        \textbf{Baseline} & \textbf{Accuracy} & \textbf{Precision} & \textbf{Recall} & \textbf{F-score} \\ \hline
        BERT & 0.589 & 0.728 & 0.589 & 0.554 \\
        T5 & \textbf{0.718} & \textbf{0.735} & \textbf{0.718} & \textbf{0.713} \\
        GPT-3.5$_{juris5}$ & 0.699 & 0.710 & 0.699 & 0.697 \\
        GPT-4$_{juris2}$ & 0.675 & 0.713 & 0.675 & 0.663 \\ \hline\hline
        \textbf{Human} & 0.812 & 0.807 & 0.812 & 0.810 \\ \hline
    \end{tabular}
    }
    \caption{Evaluation results obtained by baseline models and human predictions when the labels ``wins'' and ``partly wins'' are aggregated. \textit{Human} refers to the averaged scores of the outcome labels predicted by two human experts. For GPT-3.5 and GPT-4, the variants that achieved the highest F-scores across relevant settings are presented.}
    \label{tab:aggregated_results}
\end{table}

\textbf{Ablation study}. This paper explores a fine-grained prediction setting that differentiates between cases where the claimant wins outright and those where the claimant partially wins. This distinction inherently creates a more challenging prediction task, as accurately predicting \textit{partly wins} requires a nuanced assessment of the claimant's initial claims and the most likely outcomes for each individual claim. To understand the added difficulty of our setting, we aggregated the judgments with outcomes of ``wins'' and ``partly wins'' (\ie, treating both labels as ``wins'') and evaluated performance under this simplified setting. The overall evaluation results are presented in Table \ref{tab:aggregated_results}.\footnote{We also present results for individual categories (\ie, ``wins'', ``loses'' and ``other'') in Appendix \ref{app:partly_wins_aggregated}.} A comparison of the results in Table \ref{tab:results} and Table \ref{tab:aggregated_results} shows that all baseline models exhibit consistent improvements in prediction performance across all metrics in the simplified setting, with T5 achieving the best overall performance. Human predictions achieved a precision of 0.807 and a recall of 0.812, indicating that human annotators can effectively predict case outcomes when there is no requirement to further distinguish between the two winning-related categories.

\section{Further Discussions}
\label{sec:discussions}

\subsection{Relevance of Scores}


We would like to emphasise that the evaluation scores reported for the CLC-UKET$_{pred}$ prediction task are baseline results. Both the transformer-based and the LLM-based models could be improved further for the task at hand. For example, the latter could be further enhanced by incorporating retrieval-augmented generation \citep{lewis2020retrieval,gao2023retrieval} or chain-of-thought \citep{wei2022chain,diao2023active,kim2023cot}. Similarly, human experts might achieve better predictions by investing more time and conducting further research. Those interested in the legal domain are, therefore, encouraged to apply caution when drawing conclusions for legal practice.

The prediction task has been designed from the perspective of the claimant. This perspective informs the outcomes ``claimant wins'', ``claimant loses'', ``claimant partly wins'' and ``other''. This approach makes sense as it is first for the claimant to decide whether they apply for a decision of the Tribunal. Once the claimant has taken this first step, it is for the defendant to decide how they react to the claim. Whilst the outcome prediction for the claimant is also of relevance for the defendant, it should be noted that both models and human predictors achieve different scores depending on whether ``wins'' or ``loses'' is predicted.

Against this background, it is worth discussing a few patterns in the scores. First, both models and legal experts achieve higher recall than precision scores for ``wins'' and higher precision scores than recall scores for ``loses''. Precision is a useful measure when the costs of a wrongly predicted positive are high. In a litigation context, this is the case when the costs of initiating litigation (\eg, fees for legal and other advisers, court fees, time and stress involved) are high. Likewise, recall is a useful measure if the costs of missing a true positive are high. In the context of court proceedings, this is the case when the opportunity cost of not initiating likely successful litigation is high, for example, if the expected remedy has a high monetary value or otherwise has a high relevance for the potential claimant (\eg, for emotional reasons). Hence, it depends on the specific situation of a potential claimant whether precision or recall provides better guidance. Since the UKET currently does not charge fees and claimants can represent themselves (thereby saving costs), \textit{recall} may be the preferable score if the claim matters to the potential claimant. Second, it is worth noting that the F-score of GPT-4$_{juris2}$ for ``partly wins'' outperforms the human predictors. This may indicate the LLM's ability to navigate more complex litigation, which involves multiple claims or multiple parties on either side.

\subsection{Possible Reasons for Errors}
Models and annotators, based on the extracted facts and claims, cannot always determine whether a tribunal's decision will finally resolve the claim or only address a preliminary issue. For example, in a disability discrimination case, the tribunal might first issue a judgment confirming the claimant's disability (preliminary issue), followed by a second judgment addressing the actual discrimination claim. The first judgment (which the claimant may win) is a necessary step but does not resolve the final claim, whilst the second judgment might conclude that there was no discrimination (such that the claimant ultimately loses). Preliminary issues are often contested, and some applications may solely seek a tribunal declaration on such issues (\eg, confirming the claimant is an employee or disabled). The possibility of such multi-step proceedings increases the complexity of predictions and has likely had a negative effect on the scores of both the models and the human predictors.


Further difficulties arise in cases where the UKET renders a procedural decision instead of deciding on the substance of the claim. Such cases are classified as ``other''. However, both models and human annotators may predict a substantive instead of a procedural decision and, therefore, suggest ``claimant wins'' or ``claimant loses''. According to our annotation guidelines, this affects, in particular, the categories of ``claimant partly wins'' and ``other''. This complexity may have contributed to low evaluation scores for ``claimant partly wins'' and ``other''.

More generally, the extracted facts, which are the basis for both the models' and the humans' predictions, may not include all the elements needed to form a prediction. This may be the result of GPT-4 not including all details in the facts section when extracting the facts from the underlying UKET judgments. For example, when there is an application for costs, which is highly dependent on the parties' behaviour, the models and legal experts may be limited in their prediction due to factual details missing. Additionally, certain outcomes may hinge on factors like the respondent's failure to challenge the claim or produce evidence, which might not be reflected in the extracted facts, leading to incorrect predictions. Although extracted facts may include procedural aspects, they do not always capture procedural facts that determine the outcome, such as the timing of a claim that is dismissed due to late submissions.

\section{Conclusion}

This paper explores the prediction of dispute outcomes for the UK Employment Tribunal (UKET). It also illustrates the utility of LLMs for automatic annotation to reduce the burden of extensive manual annotation. With LLM-aided annotation, we curated the CLC-UKET dataset with comprehensive, high-quality legal annotations. We showcased how the CLC-UKET data can be used to construct a prediction task to categorise case outcomes based on sequences of facts and claims. We fine-tuned and evaluated two widely used Transformer-based models on this prediction task. We also evaluated LLMs on the prediction task with a range of settings, and reported human performance on the task to facilitate model calibration. These empirical efforts serve as a useful benchmark for the UKET prediction task. We will make the CLC-UKET dataset publicly available\footnote{The CLC website: https://www.cst.cam.ac.uk/research/srg/\\
projects/law.} to facilitate future research in this field. 


\section*{Ethics Statement}

The curated dataset is developed on the basis of the Cambridge Law Corpus (CLC), which aggregates publicly available UK legal judgments. Both the decisions in the CLC and the jurisdiction codes of UKET are licensed for use under the Open Government Licence. This licence grants a worldwide, royalty-free, perpetual and non-exclusive licence. Access to the CLC is restricted to researchers with confirmed ethical clearance and requires compliance with the DPA and UK GDPR. Whilst UK legal judgments are not anonymised, Rule 50 of the Employment Tribunal Rules ensures that sensitive personal information is anonymised when necessary. Additionally, Schedule 2, Part 5 of the DPA provides derogations for academic research, alleviating the burden of notifying all individuals involved in judgments.

Our dataset does not go beyond publicly available information and includes established procedures for data removal if requested. Like the original CLC, access to the dataset created for this paper is limited to qualified researchers who adhere to the relevant ethical and legal standards. Given the public availability of the data and our efforts to democratise access to legal information, we believe that we meet the ethical requirements for this research. 

For more details on the legal and ethical considerations concerning the underlying CLC dataset, see \citet{ostling2023cambridge}.

\section*{Limitations}

Whilst our study provides valuable insights into the prediction of dispute outcomes for the UK Employment Tribunal, it is important to acknowledge certain limitations of our findings.

\textbf{Access to the \textit{actual} facts and claims of the cases.} The facts and claims used in this paper were extracted from tribunal decisions. This was necessary given the impossibility of obtaining actual facts and claims in the number necessary for this paper. Consequently, we employed the extracted facts and claims from the court judgments as a practical substitute, providing a tangible foundation for our judgment prediction models. 

This approach could potentially introduce information biases at the input stage of the prediction task. The facts and claims that we used in the CLC-UKET dataset were derived from the judges' written decisions at the end of the proceedings. Since the judges know the result of the case at this stage of the process, the texts they write may inherently contain biased information \citep{sargeant2024}. For example, sentiment words in the judges' statements might implicitly reveal their inclinations towards certain decisions. The models might incorporate such factors when making predictions related to case outcomes. Similarly, the legal experts may have picked up such sentiments.

In subsequent research, we will explore alternative methods of identifying facts and claims to better approximate the original submissions to the court, thus fostering a more realistic modelling of judgment prediction.

\textbf{Automatic information extraction.} Manual annotation of legal texts requires extensive expert knowledge and can be costly. To alleviate these challenges, this research utilised GPT-4 for automatic information extraction. Whilst the use of GPT-4 offers notable advantages in terms of time and cost efficiency, and the extraction results are generally satisfactory according to the quality check conducted by legal experts in a related study \citep{defaria2024automatic}, this annotation practice is not without flaws. The quality of legal annotations could be further improved in future explorations. There is also room to explore the effect of extracting and providing more detailed facts compared to the relatively concise fact statements present in the current CLC-UKET dataset.

\textbf{Dataset and evolution of law over time.} We do not know whether the datasets employed are representative or include all decisions by the UKET in the relevant period. The dataset providing the cases to be predicted by the models and human experts covers the years 2011 to 2023. During this time, both employment and procedural law has evolved. Predicting a case outcome without knowing the precise decision date may lead to mistakes. Models and human predictors did not have direct access to the date at which the underlying case was decided. However, they may have inferred the decision date from the case identifier, which contains the year of the decision.

\section*{Acknowledgements}
This project received funding support from the Cambridge Centre for Data-Driven Discovery and Accelerate Programme for Scientific Discovery, made possible by a donation from Schmidt Futures. We are grateful for the helpful comments from Ludwig Bull, Mateja Jamnik, Leif Jonsson, Chikako Kanki, Måns Magnusson, Holli Sargeant, Takenobu Tokunaga, Arvid Wenestam and the participants of the ``Artificial Intelligence and Law'' workshop at Hitotsubashi University in December 2023.

\bibliography{anthology-2022,technical,legal}

\clearpage
\appendix

\section{Implementation Details}

\subsection{Experiment Settings for Transformer Models}
\label{app:experiment_details}

The implementation of the two Transformer-based models is based on the HuggingFace transformer library \citep{wolf2020transformers_huggingface}. We used the \textit{base} versions for both models, initialised from their pre-trained weights. The BERT-base checkpoint has 110 million parameters. The T5-base checkpoint has 220 million parameters. The maximum input sequence length was set to 512 tokens\footnote{All input texts to BERT and T5 are under this token length limit.}. We tried different settings for other hyperparameters such as \textit{weight decay} and the number of \textit{warm-up steps}, and found that the values of those hyperparameters have an impact on how fast the model is trained, especially at the beginning steps, but do not have a strong impact on the final learning performance. For this reason, we set both \textit{weight decay} and \textit{warm-up steps} to 0 for ease of model implementation and future replication. All training processes were performed on an Nvidia RTX 8000 GPU. 

\subsection{Final Prompts Used in the GPT-based Experiments}
\label{app:prompts}

We experimented with a number of prompts whilst exploring automatic legal annotation using GPT-4 and the prediction of case outcomes with GPT-3.5 and GPT-4. The final prompts that we used were selected based on the quality of the responses from GPT models for the task at hand. 

The \textbf{information extraction} prompt that we used to extract data from UKET court decisions reads: 

\begin{quote}
    You are a legal assistant. Your task is to read through the court decisions that I will send you, and extract the following information for each input: 1. facts of the case; 2. claims made; 3. any references to legal statutes, acts, regulations, provisions and rules, including the specific number(s), section(s) and article(s) of each of them, and including procedural tribunal rules; 4. references to precedents and other court decisions; 5. general case outcome; 6. general case outcome summarised using one of the four labels - `claimant wins', `claimant loses', `claimant partly wins' and `other'; 7. detailed order and remedies; 8. essential reasons for the decision (procedural and substantive). If there are multiple claimants or respondents, extract the case outcome for each and all of the claimants or respondents separately. Please stick strictly to the text contents that I will send.
\end{quote}

The \textbf{zero-shot} prompt that we used for the GPT-3.5 and GPT-4 prediction experiments is:

\begin{quote}
    You are a legal assistant. Your task is to predict the most likely outcome for a case based on the facts and claims that I will send you. Please summarise the case outcome using one of the four labels - `claimant wins', `claimant loses', `claimant partly wins' and `other'. Note that the label `other' is to be reserved for cases for which the result cannot be predicted or where the outcome cannot be described in terms of winning or losing (\eg, a merely procedural decision such as a stay or an evidence collection). The output should be one of the four labels.   
\end{quote}

The \textbf{few-shot} prompt that we used for GPT-3.5 and GPT-4 prediction experiments is:

\begin{quote}
    You are a legal assistant. Your task is to read through a few examples of legal case outcome prediction that I will send you and predict the most likely outcome for a case based on the facts and claims that I will send you. Please summarise the case outcome using one of the four labels - `claimant wins', `claimant loses', `claimant partly wins' and `other'. Note that the label `other' is to be reserved for cases for which the result cannot be predicted or where the outcome cannot be described in terms of winning or losing (\eg, a merely procedural decision such as a stay or an evidence collection). The output should be one of the four labels.

    To give you a few examples: 

    Case example \#1
    
    Facts: <FACTS>

    Claims: <CLAIMS>

    The case outcome label is: <OUTCOME LABEL>

    <OTHER CASE EXAMPLES>

    Case to be predicted:  

    Facts: <FACTS>
    
    Claims: <CLAIMS>

    What is the most likely case outcome?
\end{quote}

\section{Further Analysis of the CLC-UKET Dataset}

\begin{table*}[t]
\centering
\small
    \begin{subtable}[t]{\linewidth}
        \begin{tabular}{l|p{0.8\linewidth}}
        \hline
        \textbf{Facts} & The Claimant, Mr. B Shaw, was employed as a Business Adviser by the 2nd Respondent from 10 April 2007 until 30 April 2015. His employment then transferred under TUPE to the 3rd Respondent until he was made redundant on 30 June 2015. At the time of redundancy, the Claimant was 70 years old and had been continuously employed for 8 complete years. His rate of pay was £124 per day for a 4 day week, which is £496 per week. Both the 2nd and 3rd Respondents were insolvent. The Claimant was never employed by the 4th Respondent. \\ \hline
        \textbf{Claims} & The Claimant presented a claim for a redundancy payment to the Employment Tribunal. \\ \hline
        \textbf{Outcome} & Claimant wins \\ \hline
        \end{tabular}
    \caption{Case 3346845/2016.}
    \end{subtable}
    

    \begin{subtable}[t]{\linewidth}
    \vspace{0.1cm}
        \begin{tabular}{l|p{0.8\linewidth}}
        \hline
        \textbf{Facts} & Facts: The claimant, Mr P Soennecken, was employed by the respondent, Otis Limited, as a Lift Engineer. On 17 November 2017, he was asked to attend the M\&S store in Newbury because two passengers were trapped in a lift. He arrived at the store, parked outside and entered carrying his test tool but without his personal protective equipment (PPE) or other equipment provided by the respondent to ensure protection of health and safety when working on lifts. He proceeded to rescue the passengers from the lift by helping them to jump from the lift to the floor, which was just over 30cm from the floor level. He did not use a barrier to protect the gap between the lift and the floor. After he had completed the rescue of the passengers, the claimant returned to his van and collected his PPE and other equipment and proceeded to repair the broken lift. This resulted in the passengers complaining to M\&S about the claimant, which in turn led to M\&S complaining to the respondent. On receipt of the complaint, the respondent suspended the claimant pending an investigation carried out by Barry Sanderson. The allegations were breach of the cardinal rule by failing to use effective barriers, breaches of health and safety by failing to wear safety cap and gloves, not following correct procedures when releasing passengers from a lift car, a complaint in the manner the claimant spoke to the trapped passengers. Having reviewed the evidence and the representations made on behalf of the claimant, Mr Jenkinson concluded that Allegations 1, 2 and 3 were made out and he took the decision to dismiss the claimant summarily for gross misconduct. This was notified to him by letter dated 24 January 2018. He was given the right of appeal against the decision. He appealed by letter dated 25 January 2018 and the appeal meeting was held on 6 February conducted by Alex Lampe. Having reviewed the evidence and the representations made on behalf of the claimant, Mr Lampe upheld the decision to dismiss. \\ \hline
        \textbf{Claims} & The claimant's complaints of unfair dismissal and wrongful dismissal. \\ \hline
        \textbf{Outcome} & Claimant loses \\ \hline
        \end{tabular}
    \caption{Case 2204650/2018.}
    \end{subtable}
    \caption{Examples from the CLC-UKET$_{pred}$ dataset.}
    \label{tab:uket_examples}
\end{table*}

\subsection{Examples From the CLC-UKET Dataset}

Table \ref{tab:uket_examples} presents facts, claims and general case outcomes for two cases in the CLC-UKET$_{pred}$ dataset. Facts and claims are extracted annotations from GPT-4. Facts and claims are concatenated to form the input to the prediction task. Outcome labels are manually extracted by a legal expert from court judgments and are used as the target output of the prediction task.

\subsection{Page Count Distribution}

We calculated the page counts for the 52,339 court decisions in the original UKET subset in the CLC, which gives an essential idea of the length distribution of case decisions heard by the UKET. 

\begin{table}[t]
    \centering
    \resizebox{0.9\columnwidth}{!}{
    \begin{tabular}{c|c||c|c}
    \hline
        \textbf{Page count} & \textbf{\#Cases} & \textbf{Page count} & \textbf{\#Cases} \\ \hline
        1 & 32,853 & 6 & 523 \\
        2 & 8,604 & 7 & 415 \\
        3 & 1,722 & 8 & 461 \\
        4 & 1,137 & 9 & 379 \\
        5 & 722 & $\geq$ 10 & 5,523 \\\hline
    \end{tabular}
    }
    \caption{Page count distribution of the 52,339 UKET cases in the CLC.}
    \label{tab:uket_page_count}
\end{table}

From Table \ref{tab:uket_page_count}, it can be observed that the majority of cases (approximately 62.8\%) have a decision document consisting of just one page. Of these, many only contain short decisions due to procedural aspects, such as claimants withdrawing their claims or respondents not responding at all. In such instances, the court judgments do not provide substantial information on the actual facts and substantive reasoning. Against this background, we excluded most of these very brief cases when constructing the CLC-UKET dataset. 

\subsection{Jurisdiction Codes}
\label{app:juris_codes}
There are 54 jurisdiction codes linked to UKET cases\footnote{These codes are available at the UKET website at https://www.gov.uk/employment-tribunal-decisions.}. A case can be associated with multiple codes if it involves multiple issues.

Here is a comprehensive list of jurisdiction codes in UKET: employment-agencies-act-1973, rights-on-insolvency, statutory-discipline-and-grievance-procedures, religion-or-belief-discrimination, interim-relief, race-discrimination, time-to-train, notice-appeal, fixed-term-regulations, trade-union-membership, agency-workers, national-minimum-wage, written-statements, flexible-working, parental-and-maternity-leave, redundancy, harassment, human-rights, reorganisation, health-safety, unfair-dismissal, protective-award, victimisation-discrimination, written-pay-statement, maternity-and-pregnancy-rights, unlawful-deduction-from-wages, contract-of-employment, part-time-workers, sex-discrimination, equal-pay-act, disability-discrimination, practice-and-procedure-issues, public-interest-disclosure, right-to-be-accompanied, blacklisting-regulations, tax, sexual-orientation-discrimination-transexualism, time-limits, breach-of-contract, trade-union-rights, age-discrimination, certification-officer, pension, jurisdictional-points, temporary-employment, transfer-of-undertakings, working-time-regulations, renumeration, improvement-notice, european-material, time-off, reserved-forces-act, central-arbitration-committee-cac, national-security. 

\section{Aggregating ``wins'' and ``partly wins''}
\label{app:partly_wins_aggregated}

\begin{table}[t]
    \centering
    \resizebox{\linewidth}{!}{
    \begin{tabular}{c|c|ccc}
    \hline
         \textbf{Baseline} & \textbf{Label} & \textbf{Precision} & \textbf{Recall} & \textbf{F-score} \\ \hline
         \multirow{4}{*}{BERT} & wins & 0.620 & 0.902 & 0.735 \\
         & loses & 0.869 & 0.215 & 0.345 \\     
         & other & 0.036 & 0.455 & 0.067 \\
         \hline
         \multirow{4}{*}{T5} & wins & 0.819 & 0.621 & 0.707 \\
         & loses & 0.647 & 0.846 & 0.734 \\
         & other & 0 & 0 & 0 \\ \hline
         \multirow{4}{*}{GPT-3.5$_{juris5}$} & wins & 0.710 & 0.814 & 0.758 \\
         & loses & 0.720 & 0.565 & 0.633 \\
         & other & 0.143 & 0.455 & 0.217 \\ \hline
         \multirow{4}{*}{GPT-4$_{juris2}$} & wins & 0.668 & 0.883 & 0.761 \\
         & loses & 0.778 & 0.430 & 0.554 \\
         & other & 0.082 & 0.364 & 0.133 \\ \hline\hline 
         \multirow{4}{*}{\textbf{Human}} & wins & 0.832 & 0.823 & 0.828 \\
         & loses & 0.792 & 0.812 & 0.802 \\
         & other & 0 & 0 & 0 \\
         \hline
    \end{tabular}
    }
    \caption{Evaluation scores obtained by baseline models and human predictions for the three label categories when ``wins'' and ``partly wins'' are combined into a single category ``wins''. As such, \textit{wins} refers to the aggregated labels ``claimant wins'' and ``claimant partly wins''. \textit{Loses} and \textit{other} refer to the labels ``claimant loses'' and ``other'', respectively. For GPT-3.5 and GPT-4, the variants that achieved the highest F-scores across relevant settings are presented.}
    \label{tab:aggregated_individual_scores}
\end{table}

In Table \ref{tab:aggregated_individual_scores}, we present evaluation results for individual categories (\ie, ``wins'', ``loses'' and ``other'') in the ablation study where the ``wins'' and ``partly wins'' labels are aggregated. The results show that when we no longer differentiate between ``wins'' and ``partly wins'', both the baseline models and human predictions achieve higher scores for the ``wins'' category.

\section{Human Prediction for UKET Case Outcomes}
\label{app:anno_guidelines}
  
\subsection{Annotation guidelines}
\subsubsection{Introduction}
  
This UKET prediction project explores the intersection of technological innovation and access to law by predicting dispute outcomes in the UK Employment Tribunal (UKET). We implement a range of deep learning models as baselines for this task. To calibrate model performance, we are interested in investigating how well legal experts in the relevant field can predict the most likely outcomes given facts and claims of UKET cases. This investigation is of paramount importance as the human annotations can be used as a performance ``upper bound'' to facilitate more informative model comparison.  
  
\subsubsection{Data annotation}
  
Each row in the distributed data sheet corresponds to a UKET case. The information provided for the case includes the case identifier, facts of the case and claims made by the applicant(s). Annotators are asked to predict the most likely case outcome based on the facts and claims.  
  
We have 1,371 cases to be annotated in total. Case assignments: 
\begin{itemize}
    \item Annotator A: rows 2 to 1372 (1,371 cases) 
    \item Annotator B: rows 2 to 1372 (1,371 cases) 
\end{itemize}
  
\subsubsection{Annotation instructions}
  
Annotators' prediction for a case outcome should be one of the following four labels: ``Claimant Wins'', ``Claimant Loses'', ``Claimant Partly Wins'' and ``Other''. Please use the dropdown menu under the ``Annotator's Prediction (dropdown)'' column to select your predicted case outcome label.

Cases should be annotated from the perspective of the Claimant, identified as such in the Facts section. By way of example, if the claim is withdrawn, the Claimant loses because the claim is not successful. In cases where there is an Appellant and a Respondent, the Appellant is to be treated as Claimant.

The label ``Other'' is to be reserved for cases for which the result cannot be predicted (in the sense that the litigation is not about winning or losing; this does not cover uncertainty on the annotator's side) or where the outcome cannot be described in terms of winning or losing (\eg, instead of the final decision applied for, the court makes merely procedural decision such as a stay or an evidence collection). To be precise: if the Claimant applies for a procedural decision and the court awards it (does not award it), then the correct label is ``Claimant Wins'' (``Claimant Loses''). If the Claimant applies for a substantive decision (\eg, payment) and the court makes a procedural decision, which does not finally resolve the substantive application (\eg, by striking out an application for lack of jurisdiction), then the correct label is ``Other''. 

The label ``Claimant Partly Wins'' can be used when there is just one claim made or when multiple claims are made. If only one claim is made, the label ``Claimant Partly Wins'' applies if the Claimant will generally win, however, not be successful with the entirety of the claim. This is the case, where a Claimant applies for damages of £100 but will likely only be awarded £50. Additionally, you may infer a ``Claimant Partly Wins'' from other information in the Facts and Claims section than amounts. If multiple claims are made, the label applied if the Claimant will likely be successful with at least one claim in part but not with all claims in full. This is the case, where a Claimant applies for payment of wages of £100 and damages of £100 and will likely only be awarded £100 wages (but no damages). Again, a ``Claimant Partly Wins'' label may be inferred from other information than amounts. If there are multiple claims or decisions combining an outcome of ``Claimant Wins'', ``Claimant Loses'' or ``Claimant Partly Wins'' with an outcome of ``Other'' the latter shall be ignored and the case overall is to be annotated as ``Claimant Wins'', ``Claimant Loses'' or ``Claimant Partly Wins''. 

Please make predictions ONLY based on the facts and the claims. Please do not search for the case on the internet. You may consult general legal information (textbooks, internet databases, etc.) that do not refer to the specific case at hand. 

For cases where the annotators are not confident about a prediction (defined as a confidence level below 50\%), please still make a prediction using one of the four labels AND tick ``Yes'' in the ``Low Confidence'' column. This may be the case, for example, where there are only few facts or facts presented as claims the Claimant raises. Please leave the ``Low Confidence'' cell blank for cases where annotators are relatively confident about the predictions (\ie, with a confidence level greater or equal than 50\%).  
Please note down questions and comments that you may have whilst annotating the cases in the ``Notes (if any)'' column, especially if a case is complicated and hard to predict an outcome for, or if a case is interesting from the legal perspective and would be a good example for later case study. For example, it might occur that the facts section is absolutely insufficient to predict the label, in which case you should write ``insufficient facts'' in the ``Notes (if any)'' column. If there are multiple claims, and you are not confident only with regard to one of the claims, please indicate that the insufficient facts or the particular issue relate to one (and please state which one) particular claim, in the ``Notes (if any)'' section. 

Annotators should make their judgments separately (\ie, without discussions amongst themselves). This is crucial to ensure the robustness of the annotation results.  

\begin{table}[t]
    \centering
    \begin{tabular}{c|c}
    \hline
        \textbf{Label} & \textbf{Kappa Score} \\ \hline
        claimant wins & 0.322 \\
        claimant loses & 0.191 \\
        claimant partly wins & 0.284 \\
        other & 0.470 \\ \hline
    \end{tabular}
    \caption{Annotators' agreement across four label categories, measured by Cohen's Kappa scores.}
    \label{tab:kappa_scores_categories}
\end{table}

\section{Annotators' agreement across label categories}
In Table \ref{tab:kappa_scores_categories} we report Cohen's Kappa scores for the predictions of two annotators under four label categories - \textit{claimant wins}, \textit{claimant loses}, \textit{claimant partly wins} and \textit{other}. 

\end{document}